\documentclass[runningheads]{llncs}

\usepackage{eccv}

\usepackage{eccvabbrv}
\usepackage{hyperref} 
\usepackage{multirow}
\usepackage{graphicx}
\usepackage{booktabs}
\usepackage{array}
\usepackage{xcolor}
\usepackage{marvosym}
\usepackage[accsupp]{axessibility}  

\usepackage{hyperref}

\usepackage{orcidlink}

\newcommand{\corrauth}{\textsuperscript{\Letter}}

\begin{document}

\title{RECO: Region-Aware Compensation for Extrinsic Perturbations in Roadside 3D Detection} 

\titlerunning{RECO: Region-Aware Extrinsic Compensation for Roadside 3D Detection}

\author{Junsheng Du\inst{1} \and Zhaocheng He\inst{1}\textsuperscript{(\Letter)} \and
Yuhuan Lu\inst{2}\textsuperscript{(\Letter)} 
}
\authorrunning{J. Du et al.}

\institute{
School of Intelligent Systems Engineering, Sun Yat-sen University, Shenzhen, China 
\email{dujsh3@mail2.sysu.edu.cn, hezhch@mail.sysu.edu.cn}
\and
Faculty of Applied Sciences, Macao Polytechnic University, Macao SAR, China
\email{yhlu@mpu.edu.mo}
}

\maketitle

\begingroup
\renewcommand{\thefootnote}{}
\footnotetext{
\corrauth~Corresponding author.
}
\endgroup

\begin{abstract}

In intelligent transportation systems, roadside 3D object detection provides wide-area perception crucial for traffic understanding, cooperative early warning, and safe autonomous driving.
However, existing methods suffer from high sensitivity to camera extrinsics; even slight deviations (whether manifesting as transient jitter or persistent drift) can be significantly amplified by projective geometry. This cascade results in severe feature misalignment and degraded localization.
To mitigate this limitation, we propose RECO, a region-aware extrinsic compensation framework that corrects extrinsics using piecewise 6-DoF pose offsets.
RECO predicts a learnable range boundary to partition the scene into near and far regions, estimating region-specific pose corrections.
A differentiable sigmoid gate then smoothly blends the two compensated geometries to preserve continuous BEV sampling and facilitate stable optimization.
To supervise the refinement of extrinsics, we introduce an auxiliary reprojection loss that compares 2D bounding boxes projected from 3D ground truth against 2D annotations, optimizing it jointly with the standard detection objective.
Extensive experiments on the DAIR-V2X-I and Rope3D benchmarks under extrinsic perturbations demonstrate consistent improvements over state-of-the-art baselines across both yaw and $z$-axis deviations.
RECO also generalizes from transient perturbations to persistent shifts, maintaining highly competitive performance under strict calibration uncertainty.
\end{abstract}

\section{Introduction}
\label{sec:intro}
Roadside cameras provide a global viewpoint that covers intersections and long road segments well beyond the sensing horizon of on-board sensors, forming the backbone of applications such as early warning and autonomous driving systems \cite{ji2024toward, vuong2024toward}.
However, vision-based roadside 3D object detection critically depends on accurate and static camera extrinsics. Small pose perturbations can be magnified by the geometric projection to the ground plane, leading to severe feature and spatial misalignment, a phenomenon that drastically worsens at long range.

Recent works \cite{yang2023bevheight, yang2025bevheight, wang2024bevspread, wu2024heightformer} have improved robustness to camera pose jitter by mitigating sensitivity to depth estimation and stabilizing the bird's-eye-vision (BEV) transformation. These approaches typically introduce more robust BEV construction designs for image-to-BEV lifting, such as incorporating height cues to reduce brittleness under calibration noise \cite{yang2023bevheight}. Meanwhile, other architectures focus on making BEV aggregation smoother under calibration noise by enforcing stronger spatial alignment during fusion \cite{wu2024heightformer}. Collectively, these advances suggest that perception robustness under imperfect calibration benefits from jointly optimizing geometric lifting, BEV pooling, and alignment-aware fusion.
\begin{figure}[t]
\centering
\includegraphics[width=\linewidth]{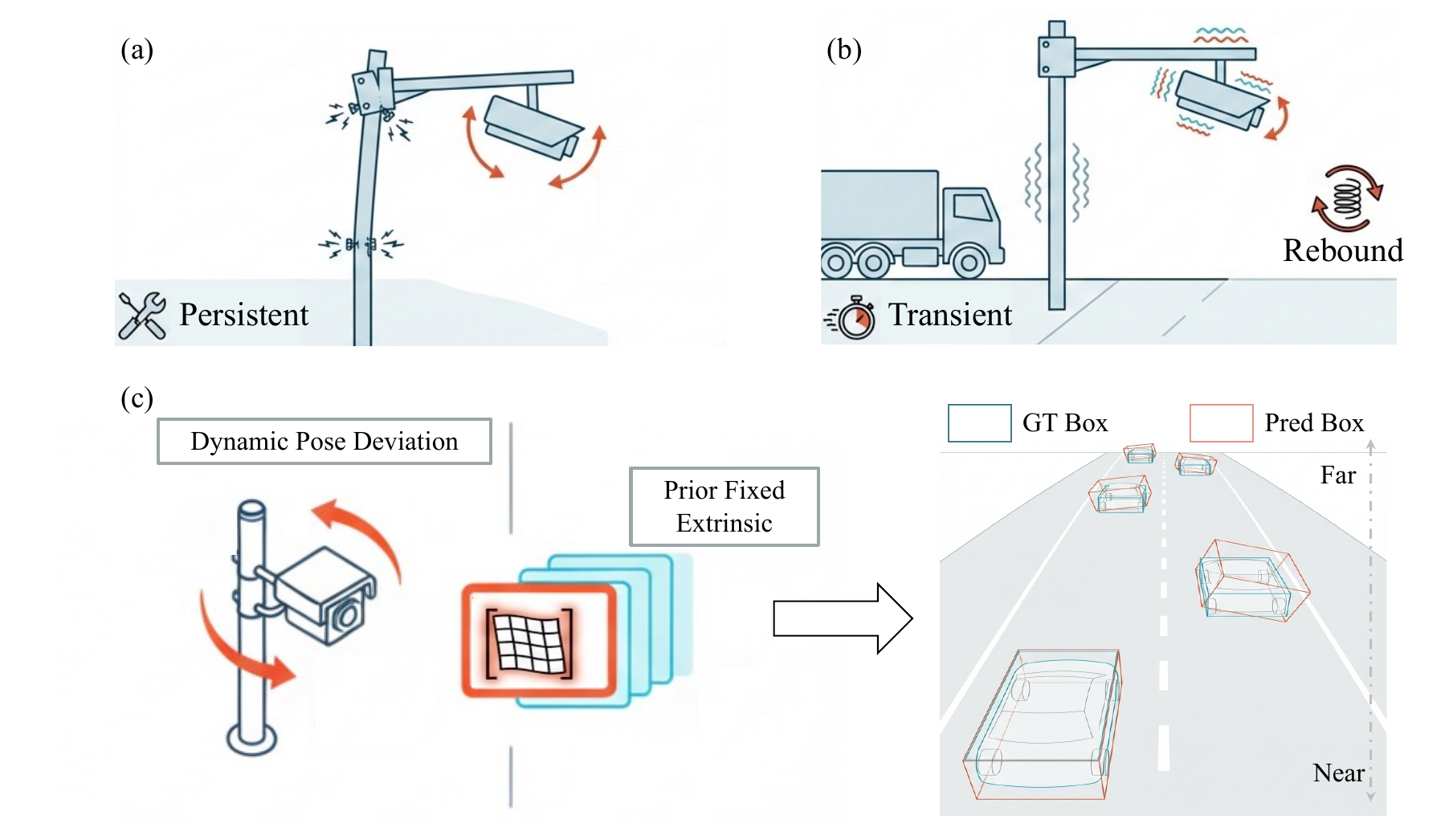}
\caption{Illustration of camera pose deviations and their impact on roadside 3D detection. (a) Persistent pose drift caused by long-term structural changes. (b) Transient pose perturbations induced by external disturbances (e.g., vibrations) that may rebound to the nominal pose. (c) Using a fixed, pre-calibrated extrinsic pose under drift causes misaligned 3D predictions, with errors typically scaling proportionally with object distance.}
\label{fig1}
\end{figure}
\

However, the aforementioned methods typically assume that camera extrinsics are fixed parameters. In real-world deployments, roadside camera poses are rarely static: thermal expansion, structural deformation, and sensor remounting can introduce subtle yet persistent drifts (Fig.~\ref{fig1}(a)). 
Moreover, transient perturbations may occur under acute external disturbances (e.g., vibrations induced by heavy commercial vehicles) and subsequently rebound to the nominal pose shortly after (Fig.~\ref{fig1}(b)). Under such dynamic pose variations, projection relying on a stale prior extrinsic inevitably induces spatial misalignment and degrades overall 3D detection performance (Fig.~\ref{fig1}(c)).

Conversely, calibration-free methods aim to enable roadside perception without explicit extrinsic priors. 
Building on this, representative approaches achieve BEV perception by learning implicit image-to-BEV transformations or incorporating geometric cues provided by V2X infrastructure \cite{10341916, s25133919}.
While effectively reducing reliance on rigorous extrinsic calibration, these approaches often compromise geometric consistency and introduce dependencies on anchor accuracy or communication reliability, which can hinder deployment scalability.

To bridge this gap and explicitly handle varying extrinsic drift, we propose RECO, a \textbf{RE}gion-based \textbf{CO}mpensation mechanism that models extrinsic refinement as piecewise 6-DoF pose offsets spanning rotation (roll, pitch, yaw) and translation ($t_x,t_y,t_z$).
RECO predicts a spatial range boundary to partition the scene into near and far regions, estimating two SE(3) corrections to refine the nominal extrinsics accordingly. To prevent spatial discontinuities induced by hard spatial assignments, we introduce a differentiable soft gate that smoothly blends the two extrinsic compensations, yielding a continuous projection field and well-behaved gradients for end-to-end training. This design yields stable BEV feature sampling despite transient camera jitter and mitigates abrupt feature artifacts caused by pose variations.

We evaluate RECO on two primary roadside 3D detection benchmarks, DAIR-V2X-I \cite{yu2022dair}, and Rope3D \cite{ye2022rope3d}, comparing it against strong BEV-based baselines. To systematically quantify robustness under controlled perturbations, we train and evaluate the models with transient extrinsic noise modeled by a zero-mean Gaussian distribution. Furthermore, we assess cross-condition generalization by applying mean-shifted extrinsic offsets during inference. Across both datasets, RECO consistently elevates detection robustness under transient perturbations and maintains strong performance when evaluated under persistent drift. In addition, ablation studies confirm that introducing the proposed compensation mechanism, particularly the projection with piecewise 6-DoF pose offsets, accounts for the majority of robustness gains and effectively neutralizes projection misalignment. In summary, our contributions are as follows:

\begin{itemize}
\item \textbf{Region-aware 6-DoF extrinsic compensation}: RECO predicts a range boundary and near/far 6-DoF pose offsets to refine nominal camera extrinsics in a spatially-adaptive, piecewise manner.
\item \textbf{Differentiable soft gating}: We introduce a sigmoid-based soft gate to smoothly interpolate near and far compensated projections, guaranteeing mathematically continuous feature fields and stable end-to-end optimization.
\item \textbf{Strong empirical results}: RECO achieves state-of-the-art robustness on DAIR-V2X-I and Rope3D, particularly under rigorously simulated extrinsic perturbations.
\end{itemize}

\section{Related Work}
We review relevant works on roadside monocular BEV-based 3D detection and 6-DoF/SE(3) extrinsic estimation, covering BEV-based designs for infrastructure perception and learning-based calibration methods.

\noindent \textbf{Roadside monocular 3D object detection}. 
Most BEV-based 3D detectors construct an image-to-BEV mapping using calibrated camera extrinsics and treat them as fixed inputs, including voxel pipelines and transformer-style BEV formulations \cite{philion2020lift, huang2021bevdet, huang2203bevdet4d, li2023bevdepth, li2024bevformer}.
For roadside monocular perception, recent works further specialize BEV construction to mitigate depth ambiguity and instability caused by discretization, improving robustness under calibration noise while still relying on nominal extrinsics \cite{yang2023bevheight, yang2025bevheight, wang2024bevspread, wu2024heightformer}.
In parallel, calibration-free approaches explore implicit image-to-BEV transformations or external geometric cues (e.g., segmentation, V2X positions) to reduce reliance on explicit calibration \cite{10341916, s25133919}.
Recent roadside monocular designs also leverage stronger priors (e.g., 2D detection prompts) to improve BEV localization and long-range stability, yet they typically do not explicitly model range-wise extrinsic drift within the projection operator \cite{ma2025pro3d}.
Overall, existing works largely improve robustness by stabilizing BEV lifting/pooling/alignment or by bypassing calibration, leaving room for approaches that adapt projection geometry online while preserving metric consistency under deployment-time extrinsic drift.

\noindent \textbf{6-DoF / SE(3) estimation}. 
A widely used strategy for handling calibration drift is to estimate a 6-DoF offset or SE(3) correction that refines nominal extrinsics from cross-modal alignment cues, instead of assuming fixed calibration. 
Early self-calibration networks regress SE(3) directly from projected depth/RGB feature or jointly enforce geometric consistency, enabling online correction under miscalibration \cite{schneider2017regnet, iyer2018calibnet, lv2021lccnet}.
More recent designs improve robustness by explicitly modeling cross-modal correspondence with cost volumes, correlation layers, or transformer-style association, followed by end-to-end regression of SE(3) updates \cite{xiao2024calibformer}.
Related to extrinsic refinement, image-to-lidar registration treats SE(3) as a relative pose between modalities and learns robust pose solvers for challenging cross-domain alignment \cite{li2021deepi2p, ren2022corri2p}.
In parallel, a large body of work studies SE(3) estimation for rigid point clouds, proposing learned correspondences and robust matching methods that are often adapted or reused for calibration refinement pipelines \cite{wang2019deep, aoki2019pointnetlk, yew2020rpm, huang2021predator, qin2022geometric, fu2021robust}.
Finally, target-free calibration methods optimize SE(3) by maximizing cross-modal consistency (e.g., mutual-information objectives), offering an alternative when supervised training signals are limited.

Despite these advances, existing BEV detectors largely treat extrinsics as fixed priors, and most SE(3) calibration methods are not integrated to handle range-dependent drift within BEV projection, motivating an extrinsic-aware, range-adaptive compensation mechanism with continuous projection for robust roadside detection.

\section{Methodology}

\begin{figure}[t]
\centering
\includegraphics[height=4.5cm]{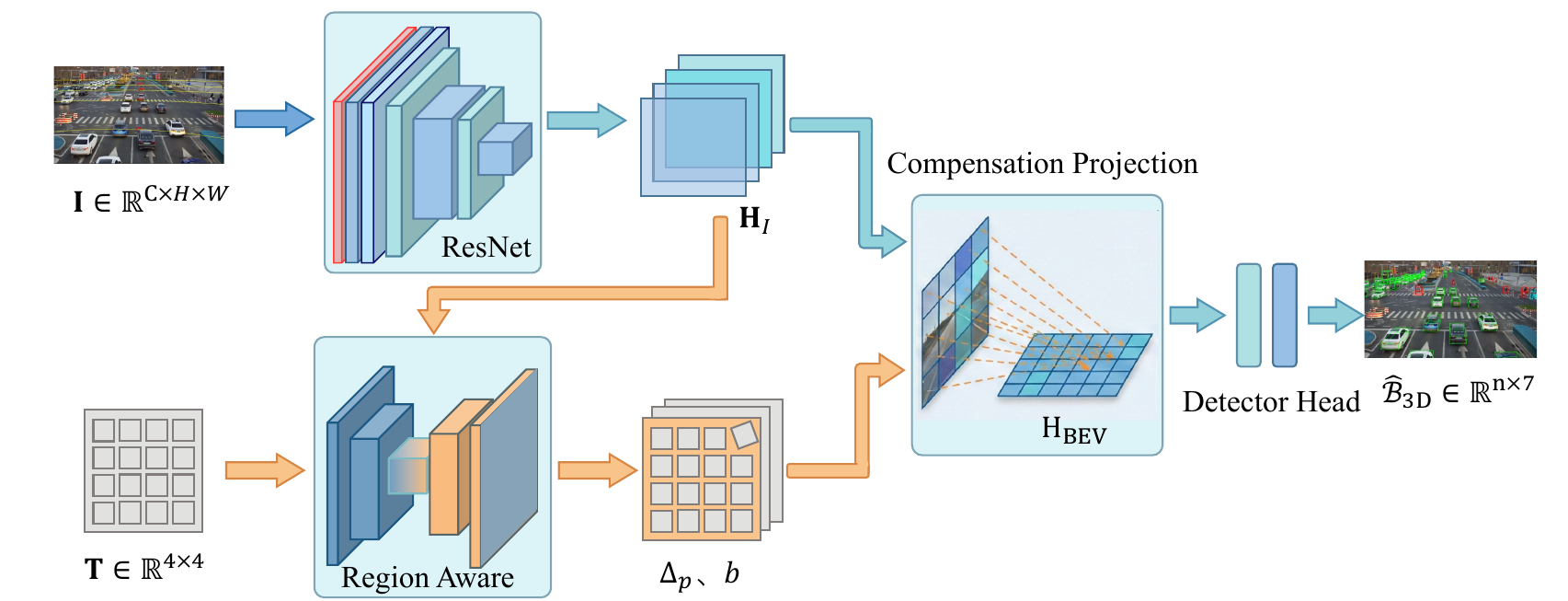}
\caption{Overview of RECO, a region-aware extrinsic compensation framework for infrastructure monocular 3D object detection. The region-aware module predicts near/far 6-DoF pose offsets $\Delta_p$ and a range boundary $b$, which are used by the compensation projection to construct BEV features for the detector head.}
\label{fig2}
\end{figure}

\subsection{Problem Formulation}
Given a single roadside RGB image $ \mathbf{I} \in \mathbb{R}^{ C \times H \times W}$ and its camera extrinsics parameterized by rotation $\mathbf{R} \in \mathbb{R}^{3 \times 3}$ and translation $t \in \mathbb{R} ^ {3}$, we denote the homogeneous transform as
\begin{equation}
      \mathbf{T} =
    \begin{bmatrix}
    \mathbf{R} & t \\
    0^\top & 1
    \end{bmatrix}
    \in \mathbb{R}^{4\times 4}
\end{equation}

Our goal is to jointly estimate a set of near/far 6-DoF pose offsets $\Delta_p \in \mathbb{R}^{1 \times 6} $, ordered as [roll, pitch, yaw, $t_x$, $t_y$, $t_z$], and a spatial boundary $b$ that partitions the scene.
Using $\Delta_p$ to compensate for camera extrinsics, the model performs 3D object detection and outputs a set of 3D bounding boxes $\hat{\mathcal{B}}_{3D} \in \mathbb{R}^{N \times 7}$, where each box is parameterized by its center $(x,y,z)$, physical size $(w, h, l)$, and heading angle $\theta$.
Formally, we seek a mapping $\mathcal{F}$ such that:
\begin{equation}
     (\Delta_p, b, \hat{\mathcal{B}}_{3D} )=\mathcal{F}(\mathbf{I}, \mathbf{T})
\end{equation}

\subsection{Overall Architecture}

The overall pipeline is illustrated in Fig.~\ref{fig2}.
Given a monocular roadside image and its nominal camera extrinsics, we first extract multi-scale 2D features using a ResNet backbone.
A region-aware extrinsic compensation network then takes the visual features alongside the prior extrinsics as input, predicting (i) 6-DoF pose offsets for near/far regions and (ii) a range boundary that partitions the scene.
Using these predicted offsets, our compensation projection module lifts image features from the perspective view onto the BEV plane to construct BEV features, which are subsequently fed into a 3D detection head to produce 3D bounding box predictions.

\subsection{Region-Aware Module}
\begin{figure}[t]
\centering
\includegraphics[height=6.5cm]{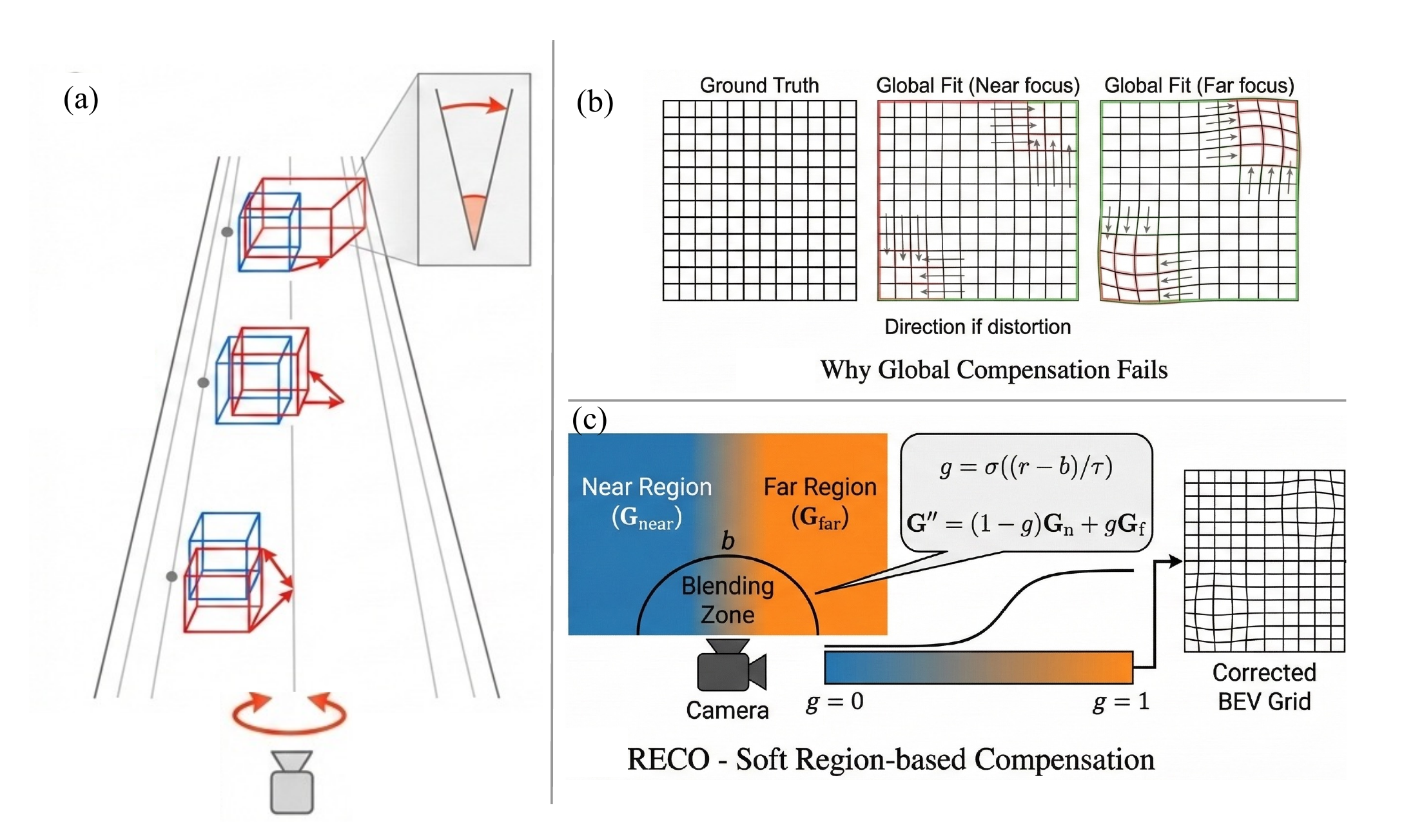}
\caption{Motivation and design for our region-aware extrinsic compensation. (a) Range-dependent effect of camera extrinsic perturbations: near-range misalignment is dominated by translation bias, while far-range misalignment is more sensitive to rotation jitter (the blue box denotes the ground-truth, while red denotes the prediction). (b) A single global 6-DoF correction cannot simultaneously fit near and far ranges under imbalanced supervision signals. (c) RECO predicts a range boundary and blends near/far SE(3) offsets.}
\label{fig3}
\end{figure}

In the roadside BEV projection, the effect of camera extrinsic perturbations is inherently range-dependent and cannot be well captured by a single global rigid correction.
In particular, near-range errors are more influenced by translational bias, while far-range misalignments are typically more sensitive to rotational jitter \cite{scott2002pose}, as illustrated in Fig.~\ref{fig3}(a).
Meanwhile, the supervision signals in the data (e.g., detection errors and reprojection errors) are not uniformly distributed across distances (Fig.~\ref{fig3}(b)).
Learning a single global compensation implicitly assumes uniform errors throughout the scene.
In practice, optimization can be dominated by a particular distance interval, hindering the learning of a globally consistent correction.

Building on the above analysis, we implement this idea with a region-aware (RA) module that jointly predicts pose corrections and range boundaries.
Given an aggregated visual feature $\mathbf{H}_I \in \mathbb{R}^{d \times H_{img} \times W_{img}}$ and nominal camera extrinsics $\mathbf{T}$, RA concatenates and encodes them into a shared latent embedding $\mathbf{Z}$. 
Here, $d$ is the dimension of the image feature.
To generalize near–far partitioning to $k$ range segments, a boundary head predicts $k-1$ boundaries via a residual form:
\begin{equation}
  b = \tilde{b} +\tanh(\mathbf{U}_b \cdot \mathbf{Z})
\end{equation}
where $ \tilde{b}$ specifies the prior boundary [$20$\,m, $40$\,m, \dots], and $\mathbf{U}_b$ is a learnable weight matrix of the boundary head.
In parallel, a pose head outputs 6-DoF offsets $\Delta_p$.
This design enables region-adaptive extrinsic compensation that aligns with the range-dependent error characteristics of BEV projection and downstream detection.

\subsection{Soft Compensation Projection Module}
\begin{figure}[t]
\centering
\includegraphics[height=4.5cm]{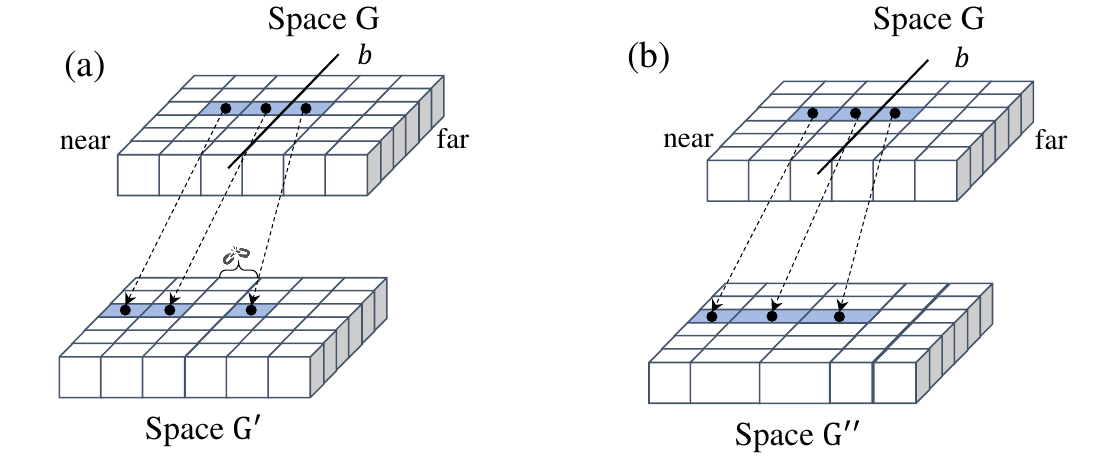}
\caption{Motivation for soft gating in range-aware extrinsic compensation. (a) Assigning piecewise offsets produces a piecewise projection, causing a discontinuous mapping at the boundary $b$ (non-smooth BEV feature sampling). (b) A differentiable soft gate smoothly blends near/far geometries, yielding a continuous projection field and stable optimization. Here, $\mathbf{G}$ denotes the original projection space. $\mathbf{G}'$ and $\mathbf{G}''$ illustrate the discontinuous/continuous projection fields after different compensations.}
\label{fig7}
\end{figure}

Using piecewise offsets can make the projection field non-smooth, introducing a discontinuity at the partition boundary (Fig.~\ref{fig7}(a)).
To address this issue, we introduce a soft compensation projection module (SCP) that enables differentiable geometry blending (Fig.~\ref{fig7}(b)).

\noindent \textbf{Range-aware geometry blending}.
Given the boundary $b$ and near/far 6-DoF offsets $\{\Delta_{p_n}, \Delta_{p_f}\}$, SCP performs a range-aware projection by smoothly blending the corresponding geometries.
We first convert $\{\Delta_{p_n}, \Delta_{p_f}\}$ to SE(3) corrections $\{\Delta\mathbf{T}_n,\Delta\mathbf{T}_f\}$ and left-compose each correction with the original extrinsic $\mathbf{T}$:
\begin{equation}
  \mathbf{P}_n=\Delta\mathbf{T}_n\mathbf{T}, \qquad
  \mathbf{P}_f=\Delta\mathbf{T}_f\mathbf{T}
\end{equation}

Given the corrected extrinsic $\mathbf{P}_*$, we construct the BEV sampling geometry by projecting each BEV cell center from the ego frame into the camera image.
Concretely, we back-project a 3D point $\mathbf{X}_e=[x,y,z,1]^\top$ in the ego frame, transform it to the camera frame via: 
\begin{equation}
 \mathbf{X}_c = \mathbf{P}_*\,\mathbf{X}_e
    = \begin{bmatrix}
    X_c\\
    Y_c\\
    Z_c\\
    1
    \end{bmatrix}.
\end{equation}

Then we apply the camera intrinsics $\mathbf{K}$ to compute the pixel position $(u,v)$ as follows:
\begin{equation}
  \mathbf{X}_k =  \begin{bmatrix}
    \tilde{x}\\
    \tilde{y}\\
    \tilde{z}
    \end{bmatrix}
    = \mathbf{K}\ \begin{bmatrix}
    X_c\\
    Y_c\\
    Z_c
    \end{bmatrix}, \qquad
 u = \tilde{x} / \tilde{z}, \qquad
 v = \tilde{y} / \tilde{z}
\end{equation}

Stacking $(u,v)$ for all BEV cells forms the geometry mapping field $\mathbf{G}_*$, which is used to sample image features and construct BEV features.

Keeping other parameters fixed, the corrected transforms $\mathbf{P}_n$ and $\mathbf{P}_f$ yield a geometry mapping field $\mathbf{G}_n$ and $\mathbf{G}_f$ for BEV projection.
For each BEV location $(x, y)$, we compute its distance $r=\sqrt{x^2 + y^2}$ in the BEV plane, then via a soft gate as:
\begin{equation}
    g=\sigma(\frac{r-b}{\tau})
\end{equation}
where $\tau$ controls the transition width. The final geometry field is obtained by gated interpolation:
\begin{equation}
    \mathbf{G}''=(1-g)\mathbf{G}_n+g\mathbf{G}_f
\end{equation}
which preserves a continuous projection field and avoids the non-smoothness introduced by hard near/far switching.

\noindent \textbf{BEV feature construction}.
We construct BEV features by warping and aggregating image features onto a discrete BEV grid.
Given backbone features $\mathbf{H}_{I}$, $\mathbf{G}''$ assigns each sample to BEV location $(x, y)$, which is quantized into BEV indices $(i,j)$ :
\begin{equation}
  i = \left\lfloor \frac{x - x_{\min}}{\Delta x} \right\rfloor, \qquad
  j = \left\lfloor \frac{y - y_{\min}}{\Delta y} \right\rfloor
\end{equation}
where $(x_{\text{min}}, x_{\text{max}}, y_{\text{min}}, y_{\text{max}})$ are the predefined BEV bounds, and $(\Delta x, \Delta y)$ are the resolution.

We then splat the projected image features onto the BEV grid and aggregate all contributions using scatter-based pooling.
The BEV feature map $\mathbf{H}_{\text{BEV}}\in\mathbb{R}^{C\times H_{\text{bev}}\times W_{\text{bev}}}$ is computed by weighted averaging:
\begin{equation}
\mathbf{H}_{\text{BEV}}[:, i, j] =
\frac{\sum \mathbf{U}_{bev}\,\mathbf{H}_I}{\sum \mathbf{U}_{bev} + \epsilon}
\end{equation}
where summations are over samples assigned to cell $(i,j)$, $\mathbf{U}_{bev}$ denotes an optional validity weight, and $\epsilon$ avoids division by zero.
Then we aggregate projected image features into the BEV grid via scatter-based pooling: features assigned to the same cell are averaged and normalized by the corresponding cell count.

\noindent \textbf{3D box prediction}.
Given the BEV feature map $\mathbf{H}_{\mathrm{bev}}$, we adopt a standard BEV detection head to produce 3D bounding box predictions.
The head first applies several BEV encoder layers to enhance spatial context, and then outputs dense regression and classification maps on the BEV grid.
Following common practice in BEV-based detectors \cite{yang2023bevheight,yang2025bevheight}, we decode the head outputs into a set of $N$ boxes
$\hat{\mathcal{B}}_{3D}$ in the ego frame, together with confidence scores.
During training, these predictions are matched to ground truth following the assignment strategy used in the baseline, and the resulting detection loss is denoted as $\mathcal{L}_{\mathrm{det}}$.

\subsection{Training Objective}

We train the detector with the standard 3D detection loss used in previous work \cite{yang2023bevheight,yang2025bevheight} and an auxiliary reprojection loss that supervises the predicted extrinsic compensation; 3D box annotations supervise the detection loss, while 2D box annotations supervise the reprojection loss.
During inference, we discard the reprojection branch and only use the predicted extrinsic compensation for geometry construction.

Specifically, given a 3D ground-truth box, we project its 3D corners to the image plane using the compensated projection to obtain a 2D box $\hat{\mathcal{B}}_{2D}$.
We then penalize the discrepancy to the corresponding ground-truth 2D box $\mathcal{B}_{2D}$ with an L2 loss over valid instances:
\begin{equation}
  \mathcal{L}_{rep}=\frac{1}{\sum m}\sum ||\hat{\mathcal{B}}_{2D} - \mathcal{B}_{2D}||^2_2
\end{equation}
where $m$ denotes the valid mask. The overall training objective is
\begin{equation}
\mathcal{L}=\mathcal{L}_{\mathrm{det}}+\lambda_{\mathrm{rep}}\mathcal{L}_{\mathrm{rep}},
\label{eq:important}
\end{equation}
where $\mathcal{L}_{det}$ is the 3D detection loss and $\lambda_{\mathrm{rep}}$ balances the reprojection supervision.

\section{Experiments}
\subsection{Experiments Setup}

\noindent \textbf{Datasets.} We evaluate RECO on two prominent benchmark datasets for roadside 3D detection: \textbf{DAIR-V2X} \cite{yu2022dair} and \textbf{Rope3D} \cite{ye2022rope3d}. \textbf{DAIR-V2X} is a large-scale benchmark for vehicle-infrastructure cooperative perception, providing synchronized multi-modal data captured at urban intersections using both vehicle-side and infrastructure-side sensors. \textbf{Rope3D} is a diverse dataset designed specifically to benchmark monocular 3D object detection from infrastructure cameras, featuring an extensive set of images collected across varied camera setups, viewpoints, and environmental conditions.

\begin{table}[t]
\centering
\caption{Comparison on the DAIR-V2X-I \cite{yu2022dair} validation set under extrinsic perturbations. Following \cite{xia2025robust}, We randomly sample angles and shifts from $\mathcal{N}(0,0.5)$ for yaw and $z$. All methods are retrained under the same protocol for a fair comparison. We report AP for car ($IoU$=0.5), pedestrian ($IoU=0.25$), and cyclist ($IoU=0.25$). "$\uparrow$" indicates that higher values are better.}
    \resizebox{\textwidth}{!}{
    \large
        \begin{tabular}{lccccccccccccc}
        \toprule
        \multirow{2}{*}{Method} & \multirow{2}{*}{Deviation} &  \multicolumn{3}{c}{Car$_{IoU=0.5}\uparrow$} & \multicolumn{3}{c}{Ped.$_{IoU=0.25}\uparrow$ }& \multicolumn{3}{c}{Cyc.$_{IoU=0.25}\uparrow$}  & \multirow{2}{*}{FPS}   \\
        \cmidrule(lr){3-5}\cmidrule(lr){6-8}\cmidrule(lr){9-11}
        & & easy & moderate & hard & easy & moderate & hard &easy & moderate & hard  \\
        \midrule  
        BEVHeight  & \multirow[c]{6}{*}{yaw} & 65.5595 &  \underline{61.7935} & \underline{61.8782} & 10.9768 &  10.3565 & 10.4573 &   39.9807 & 38.2060 & \underline{34.0681} & 19.81\\ 
        CoBEV &  & 31.4301 & 26.5978 & 26.9728 & 5.1976 &  5.0362 & 5.0865 & 23.3402 &   22.5645 & 20.7799 & 10.16 \\
        BEVSpread &  & 67.0991 & 56.0408 & 56.1161 & \underline{16.1610} & \underline{15.3372}& \underline{15.3543} & \underline{41.4683} &  \underline{41.2027}&  33.9091 & 7.78\\
        HeightFormer &  & 38.6363 & 32.0780 & 32.1450 & 9.2197& 8.8998 & 8.9994&14.9138 & 14.7369 & 18.8614 & 8.66\\
        BEVHeight++ &  & \underline{69.6915}& 58.2977 & 60.3824 & 5.7638& 5.2098 & 5.2912 & 19.4069& 18.7940 & 19.4149 & 7.80\\
        Our &  & \textbf{70.4639} & \textbf{63.0142} & \textbf{ 63.0296} & \textbf{16.2161}  &  \textbf{15.3758} & \textbf{15.4261} &  \textbf{43.3582}&  \textbf{42.9046}& \textbf{36.4710} & 17.66 \\
        \midrule
        BEVHeight  & \multirow[c]{6}{*}{z} & \underline{58.5738}& \underline{52.8735}& \underline{53.0879} & 3.4300& 3.3005&3.3102 &26.3826&25.8640& 19.1211 & 19.82\\ 
        CoBEV &  & 26.7985& 22.8999& 22.8620 & 6.7144& 6.2737 & 6.2747 &26.4755 &26.5794& 16.0429 & 10.16 \\
        BEVSpread &  & 53.9274 & 46.4839 & 46.2409 &\underline{ 11.5392}& \underline{ 10.8755 }& \underline{ 10.8755} &27.4744  &  27.1533  & 23.0330 & 10.17 \\
        HeightFormer &  & 27.6363 & 20.0780 & 20.0047 & 9.2197& 8.8998 & 8.1294 &14.8091 & 14.7369 &8.8614 & 8.67\\
        BEVHeight++ &  &55.8025 & 47.1745 & 47.1523 & 8.8375& 8.5998 & 8.7898& \underline{32.1752}& \underline{31.8063}& \underline{28.1942} & 7.79 \\
        Our &  & \textbf{72.6160} & \textbf{ 64.0757} & \textbf{64.1666} &  \textbf{13.1446} & \textbf{12.5199} &  \textbf{12.6940}&  \textbf{35.9725}&  \textbf{35.3335} & \textbf{31.5858} & 17.66 \\
        \bottomrule 
        \end{tabular}
    }
    \label{table1}
\end{table}

\noindent \textbf{Implementation details}. 
For architectural details, we apply ResNet 50 \cite{he2016deep} as the backbone network and LSSFPN \cite{philion2020lift} to construct the 3D volumetric feature. 
The model is trained with AdamW \cite{loshchilov2017decoupled} using a weight with decay of $10^{-4}$, $\lambda_{\mathrm{rep}} = 0.1$ and $\tau = 0.5$. 
The BEV grid spans [-51.2 m, 51.2 m] in width and [0 m, 140 m] in length for two datasets. For input images, we set the resolution to (864, 1536). Following \cite{li2023bevdepth, yang2025bevheight}, we apply image-level augmentations, including random cropping and rotation, as well as BEV-level augmentations, including random scaling, flipping, and rotation. All reproduction methods are trained for 50 epochs on 4 RTX A6000 GPUs with a batch size of 8. 

\begin{table}[t]
\centering
\caption{Comparison on the Rope3D \cite{ye2022rope3d} validation set under extrinsic perturbations.}
    \resizebox{\textwidth}{!}{
    \large
        \begin{tabular}{lcccccccccccc}
        \toprule
        \multirow{2}{*}{Method} & \multirow{2}{*}{Deviation} &  \multicolumn{3}{c}{Car$_{IoU=0.5}\uparrow$} & \multicolumn{3}{c}{Big Vehicle$_{IoU=0.5}\uparrow$} \\
        \cmidrule(lr){3-5}\cmidrule(lr){6-8}
        & & easy & moderate & hard & easy & moderate & hard\\
        \midrule  
        BEVHeight & \multirow[c]{4}{*}{yaw} & 43.9457& 39.1533&39.0536 & \underline{47.5926}& \underline{46.5524}&\underline{46.5961} \\ 
        BEVHeight++ &  & \underline{50.3703}& \underline{47.4298}&  \underline{45.1325} &  47.2277& 44.3472&44.3624\\
        BEVSpread &  & 41.4254 &  35.4885 & 33.2370&41.5304& 34.8029& 34.6859\\ 
        Our &  & \textbf{65.3942} & \textbf{62.3413} & \textbf{60.5443} &  \textbf{54.5881}& \textbf{56.2474}&  \textbf{56.2737}\\
        \midrule  
        BEVHeight & \multirow[c]{4}{*}{z} &46.3001& 44.7140& 44.7303& 40.4384&\underline{41.2793}&\underline{41.3240} \\ 
        BEVHeight++ &  & \underline{50.7617}& 35.2466& 32.9790 &  \underline{42.0146}&35.1928&35.0993\\
        BEVSpread &  & 41.8654&32.0237&32.0219&32.3425&32.2786&32.3853\\ 
        Our &  & \textbf{65.6319} & \textbf{60.4119} & \textbf{58.2793} &  \textbf{51.5508}& \textbf{ 52.6199}&  \textbf{52.6557}\\
        \bottomrule 
        \end{tabular}
    }
    \label{table2}
\end{table}

\subsection{Roadside 3D Detection}
\textbf{Comparison methods}. We compare different BEV-based methods, such as BEVHeight (CVPR 2023 \cite{yang2023bevheight}), CoBEV (TIP 2024 \cite{shi2024cobev}), BEVSpread (CVPR 2024 \cite{wang2024bevspread}), HeightFormer (TGRS 2024 \cite{wu2024heightformer}), BEVHeight++ (TPAMI 2025 \cite{yang2025bevheight}).

\noindent \textbf{Perturbations details}. We report robustness results under yaw rotation and $z$-axis translation, since these deviations are common in roadside deployments and induce direct range-dependent BEV misalignment \cite{zhu2025mamv2xcalib, shi2024cobev}.
Although RECO predicts full 6-DoF corrections, additional evaluations under pitch/roll perturbations are provided in the supplementary material for completeness.

\noindent \textbf{Results under transient extrinsic perturbations}. We compare RECO with other BEV-based methods like BEVHeight \cite{yang2023bevheight} and calibration-free methods like CBR \cite{10341916} on the DAIR-V2X-I val set. 
As can be seen from Table~\ref{table1}, RECO achieves the best AP for car across all splits, improving over the strongest baseline by 0.77\%, 1.22\%, and 1.15\%, and also yields consistent gains for pedestrian and cyclist.
Under $z$ perturbations, RECO surpasses the best baseline by a significant margin of 14.04\%, 11.20\%, 11.08\% on Car, and improves pedestrian by about 1.61\% to 1.82\%.
For cyclists, RECO is competitive and is notably stronger on the hard split, while CoBEV is slightly higher on easy/moderate.

On the Rope3D dataset, we reproduce the BEV-based methods listed in Table~\ref{table2} and compare our RECO against them.
Under yaw jitter, RECO exceeds the best BEV baseline by about 15\% across splits, and boosts big vehicle to 54.59\%, 56.25\%, and 56.27\%. 
Under $z$ perturbations, RECO again dominates, reaching 65.63\%, 60.41\%, 58.28\% on car, with improvements of roughly 13.55\% to 15.70\% AP over the strongest baseline depending on the split.

\begin{table}[t]
\centering
\caption{Robustness to persistent extrinsic deviations. All methods are trained with transient noise sampled from $\mathcal{N}(0,0.5)$ and evaluated AP of car ( $IoU=0.5$, easy) under persistent deviations applied to either yaw rotation or $z$-axis translation. The columns labeled $(\mu, 0.5)$ denote evaluation with a mean shift $\mu \in \{-2,-1,0,1,2\}$, while keeping the same noise scale $(0.5)$. "$\uparrow$" indicates higher is better.}
    \resizebox{\textwidth}{!}{
    \large
        \begin{tabular}{lcccccccccccc}
        \toprule
        Method & Deviation &${(-2, 0.5)}\uparrow$&${(-1, 0.5)}\uparrow$ &${(0, 0.5)}\uparrow$ & ${(1, 0.5)}\uparrow$ &${(2, 0.5)}\uparrow$ \\
        \midrule  
        BEVHeight & \multirow[c]{4}{*}{yaw} & 18.0282&56.7656&67.5595&51.4496&10.2972\\ 
        CoBEV &  &10.0174&21.8748&31.4301&27.4911&6.434\\
        HeightFormer &  & 3.3049&12.7204&38.6363&10.9978&5.4677\\ 
        BEVHeight++ &&13.7437&54.039&69.6915&52.6252&16.4732\\
        Our &  & \textbf{49.3382}&\textbf{70.0774}&\textbf{70.4639}&\textbf{70.1092}&\textbf{44.3605} \\
        \midrule  
        BEVHeight & \multirow[c]{4}{*}{z} & 5.307&16.7973&48.3313&16.196&4.6595\\ 
        CoBEV &  & 12.7873 &24.6159 &46.0432 &24.9035 &13.2877\\
        HeightFormer &  & 3.2845 &  15.1936 & 27.6363&13.7415& 2.9499\\ 
        BEVHeight++&& 26.8912&\textbf{42.5294}&55.8025&45.6132&31.2033\\
        Our &  & \textbf{31.0973}&{37.3848}&\textbf{72.616}&\textbf{64.3633}&\textbf{57.2835}\\
        \bottomrule 
        \end{tabular}
    }
    \label{table3}
\end{table}

\noindent \textbf{Results of persistent extrinsic perturbations}. Table~\ref{table3} evaluates cross-condition generalization by training with transient extrinsic jitter and testing under persistent mean-shifted deviations.
All models are trained with zero-mean Gaussian noise $\mathcal{N}(0,0.5)$ to mimic short-term disturbances and are then evaluated on car AP ($IoU=0.5$, easy) with a mean shift $\mu\in \{-2,-1,0,1,2\}$ applied to yaw rotation or $z$-axis translation.
Under this setting, our method achieves the best performance across both yaw and $z$ perturbations, demonstrating improved robustness when models trained for transient noise are deployed under persistent miscalibration.

We observe a clear asymmetry between the yaw and $z$ deviations.
For yaw rotation, the performance under positive and negative mean shifts is approximately symmetric, suggesting that the detector’s robustness is largely invariant to left–right perturbations.
In contrast, for translation along the $z$-axis, the model exhibits higher tolerance to positive shifts than to negative ones.
This behavior is physically plausible: $z$-translation changes the effective camera height and thus alters the apparent object scale and depth distribution in a direction-dependent manner, whereas yaw rotation induces a more symmetric left-right reprojection effect.

\begin{table}[hbtp]
\centering
\caption{Comparison of different rotations on the DAIR-V2X-I. }
    \resizebox{\textwidth}{!}{
    \large
        \begin{tabular}{lccccccccccccc}
        \toprule
        \multirow{2}{*}{Method} & \multirow{2}{*}{Deviation} &  \multicolumn{3}{c}{Car$_{IoU=0.5}\uparrow$} & \multicolumn{3}{c}{Ped.$_{IoU=0.25}\uparrow$ }& \multicolumn{3}{c}{Cyc.$_{IoU=0.25}\uparrow$}    \\
        \cmidrule(lr){3-5}\cmidrule(lr){6-8}\cmidrule(lr){9-11}
        & & easy & moderate & hard & easy & moderate & hard &easy & moderate & hard  \\
        \midrule  
        BEVHeight  & \multirow[c]{3}{*}{pitch} & \underline{53.1881} & \underline{46.2226} & \underline{46.3931}& \textbf{9.3655}&\textbf{9.1158}& \textbf{9.2308}& 29.3548&  \textbf{29.8336}& \textbf{25.2666}\\ 
        BEVHeight++ &  & 50.5844&  42.2217&  42.1829 & 7.6196& 7.3630& 7.3657 & 27.0724& 26.7482& 23.2941  \\
        Our &  & \textbf{65.3235} & \textbf{57.1345} & \textbf{57.9029} & \underline{8.4951}&  \underline{8.1553}& \underline{8.2599}&  \textbf{30.2731} & \underline{29.6824}&\underline{25.1420} \\
        \midrule
        BEVHeight  & \multirow[c]{3}{*}{roll} & \underline{52.3127}& \underline{46.8362}& \underline{47.0078} &{6.0710}&  5.6716&  5.708&27.0194&26.4693& 19.7767\\       
        BEVHeight++ &  &50.4606& 42.1012& 42.0723& \underline{6.7628}&   \underline{6.6048}& \underline{6.2304}&\underline{29.1563} &  \textbf{29.8950}&  \textbf{23.0539}  \\
        Our &  & \textbf{62.3741}& \textbf{54.0858}& \textbf{54.1871} & \textbf{ 6.9194}& \textbf{ 6.6530}& \textbf{6.4159}&\textbf{29.2147} &\underline{ 27.7538}& \underline{22.7853}\\
        \midrule
        BEVHeight  & \multirow[c]{3}{*}{pitch \& roll} & \underline{51.0522}& \underline{44.2283}& \underline{44.3883}&   5.7002& 5.3966&  5.4665 & 25.4191&24.9127&19.5592\\ 
        BEVHeight++ &  & 49.7541& 41.4667&  41.4342& \textbf{5.8089}& \textbf{5.6719}&\textbf{5.6707}& \underline{27.0194}&\underline{26.4693}& \underline{19.7767}\\
        Our &  & \textbf{58.8318}&  \textbf{50.8446}& \textbf{ 50.9040}& \underline{5.7964}& \underline{5.5653}& \underline{5.6622}& \textbf{27.4047}& \textbf{26.7902}&\textbf{20.8360}\\
        \bottomrule 
        \end{tabular}
    }
    \label{table5}
\end{table}

\noindent \textbf{Results on rotational robustness}. 
To evaluate robustness against different perturbations, we test all methods trained under yaw perturbation $\mathcal{N}(0, 0.5)$ on pitch, roll, and joint pitch\&roll perturbations, without retraining or fine-tuning.
Table~\ref{table5} reports the comparison results on DAIR-V2X-I.
Following BEVHeight++ \cite{yang2025bevheight}, the angles sampled from $\mathcal{N}(1,1.67)$ are applied to the camera pitch and roll at test time. 
Our method outperforms comparative methods, with gains over BEVHeight \cite{yang2023bevheight} under pitch perturbations reaching +12.14\%, +10.91\%, +11.51\% in the car category.
For the cyclist, our method also provides competitive performance, achieving 30.27\%, 29.68\%, 25.14\% AP under pitch deviation and 27.40\%, 26.79\%, 20.84\%  under joint pitch\&roll deviation, both better than BEVHeight++. 
Overall, these results show that our method provides markedly better robustness to rotational extrinsic perturbations, especially for car and cyclist detection.

\begin{figure}
\centering
\includegraphics[width=\linewidth,height=0.8\textheight,keepaspectratio]{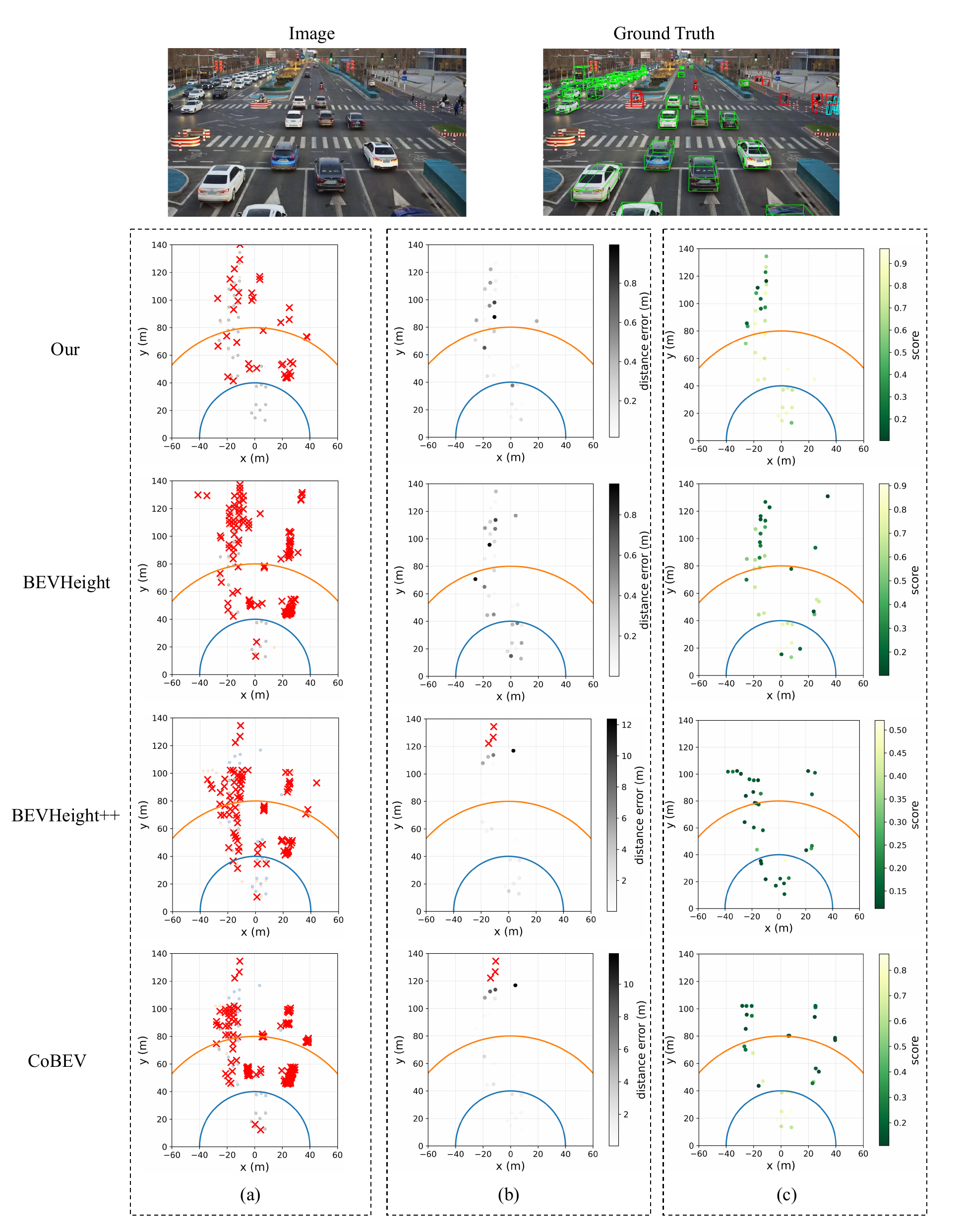}
\caption{Qualitative comparison under extrinsic perturbations. \textbf{Top}: a representative roadside image (left) and the corresponding ground-truth annotations (right). \textbf{Bottom}: BEV visualizations for different methods (rows). (a) Missed and false detections: red crosses denote missed ground-truth objects and unmatched predictions, while dots indicate ground-truth and matched predictions. (b) Distance error map for matched predictions: darker colors denote larger distance errors between predicted and ground-truth, while red crosses indicate missed ground-truth. (c) Confidence map of predictions: brighter colors indicate higher detection confidence. The blue and orange semicircles indicate range rings.} 
\label{fig5}
\end{figure}

\subsection{Ablation Study}

\begin{table}[t]
\centering
\caption{Ablation study of compensation strategies under persistent extrinsic deviations. All variants are trained under transient noise $\mathcal{N}(0, 0.5)$ and evaluated with mean-shifted deviations $(\mu, 0.5)$, where $\mu \in \{-2, -1, 0, 1, 2\}$, applied to either yaw rotation or $z$-axis translation. We compare No\_Comp, Global\_Comp, Hard\_Comp (range-aware compensation with hard assignment), and Soft\_Comp (range-aware compensation with soft gating). "$\uparrow$" indicates higher is better.}
    \resizebox{\textwidth}{!}{
    \large
        \begin{tabular}{lcccccccccccc}
        \toprule
        Method & Deviation &${(-2, 0.5)}\uparrow$&${(-1, 0.5)}\uparrow$ &${(0, 0.5)}\uparrow$ & ${(1, 0.5)}\uparrow$ &${(2, 0.5)}\uparrow$ \\
        \midrule  
        No\_Comp & \multirow[c]{4}{*}{yaw} & 18.0282&56.7656&67.5595&51.4496&10.2972\\ 
        Global\_Comp &  &32.2572&33.2531&33.2416&33.2421&33.0389\\
        Hard\_Comp &  &\underline{40.0773}&\underline{64.9268}&\underline{68.0451}&\underline{65.7611}&\underline{40.0988}\\ 
        Soft\_Comp &  & \textbf{49.3382}&\textbf{70.0774}&\textbf{70.4639}&\textbf{70.1092}&\textbf{44.3605} \\
        \midrule  
        No\_Comp & \multirow[c]{4}{*}{z} & 5.307&16.7973&48.3313&16.196&4.6595\\ 
        Global\_Comp &  &23.0018&23.2671&23.2683&23.3090&23.2668\\
        Hard\_Comp &  & \textbf{31.6210}&\underline{36.8859}&\underline{68.3201}&\textbf{65.0907}&\underline{56.7534}\\ 
        Soft\_Comp &  & \underline{31.0973}&\textbf{37.3848}&\textbf{72.616}&\underline{64.3633}&\textbf{57.2835}\\
        \bottomrule 
        \end{tabular}
    }
    \label{table4}
\end{table}

Table~\ref{table4} demonstrates the effectiveness of introducing compensation into the projection pipeline under persistent mean-shifted extrinsic deviations. 
Compared with No\_Comp, adding a Global\_Comp mechanism substantially improves robustness in several shifted settings, indicating that a single global 6-DoF correction can partially alleviate projection misalignment under persistent drift.
More importantly, range-aware compensation yields consistent additional gains over global compensation.
Both Hard\_Comp and Soft\_Comp significantly improve performance under large mean shifts, and Soft\_Comp performs best overall across yaw and $z$ perturbations.
This suggests that range-adaptive modeling is beneficial and that differentiable soft gating mitigates the discontinuities of hard partitioning, leading to stronger robustness.

\subsection{Visualization Results}

\begin{figure}[t]
\centering
\includegraphics[width=\linewidth]{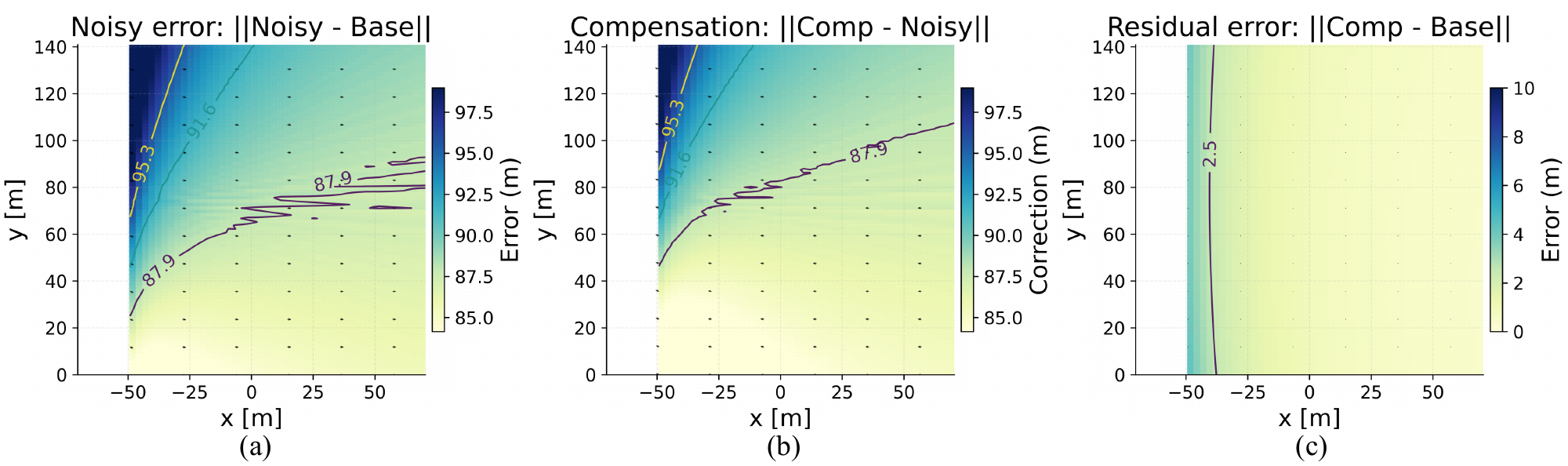}
\caption{Visualization of BEV grid displacement and compensation under yaw $\mathcal{N}(0,0.5)$. (a) Displacement magnitude after injecting extrinsic noise, measured as $||\mathbf{G}_\mathrm{noisy} - \mathbf{G}_\mathrm{base}||$ at each BEV cell. (b) The compensation applied by our method, $||\mathbf{G}_\mathrm{comp} - \mathbf{G}_\mathrm{noisy}||$, shows how the geometry field is corrected across the grid. (c) Residual displacement after compensation, $||\mathbf{G}_\mathrm{comp} - \mathbf{G}_\mathrm{base}||$. The overlaid polylines are iso-contours of the plotted quantity. Colorbars indicate the displacement/correction magnitude in meters.} 
\label{fig6}
\end{figure}

\textbf{Detection result}. Fig.~\ref{fig5} provides a qualitative comparison of RECO and prior methods under the same scene and yaw deviations $\mathcal{N}(0,0.5)$.
As shown in Fig.~\ref{fig5}(a), prior methods exhibit noticeably more false positives or missing ground-truth boxes, especially in the far-range region, while RECO yields fewer failure cases. 
The error map in Fig.~\ref{fig5}(b) further shows that RECO reduces distance errors for matched detections, with fewer high-error points in the far-range region, suggesting improved detection effectiveness under extrinsic perturbations. 
In addition, Fig.~\ref{fig5}(c) shows that RECO produces more reliable confidence estimates, assigning higher scores to correct detections.

\noindent \textbf{Compensation result}. Fig.~\ref{fig6} provides direct evidence that the proposed compensation restores the consistency of BEV sampling under extrinsic perturbations.
The noise error displacement concentrates and grows with range (Fig.~\ref{fig6}(a)), indicating a highly non-uniform distortion of the geometry field rather than a simple global shift.
Our method produces a correction that closely counteracts this distortion, yielding near-zero residual displacement over most of the BEV plane (Fig.~\ref{fig6}(b) and Fig.~\ref{fig6}(c)).
This shows that RECO effectively cancels the geometric misalignment introduced by pose noise and recovers an approximately noise-free projection geometry. 

\section{Conclusion}
In this paper, we address the long-standing sensitivity of roadside monocular 3D detection to camera extrinsic drift in real deployments. We presented RECO, a region-aware extrinsic compensation framework that predicts a learnable range boundary and estimates near/far 6-DoF pose corrections to refine nominal extrinsics. To avoid non-smooth BEV sampling induced by region-aware offsets, we further introduced a differentiable soft gated compensation projection that smoothly blends the compensated geometries and enables stable optimization with reprojection supervision. Extensive experiments on DAIR-V2X-I and Rope3D demonstrate that RECO consistently improves robustness under both transient jitter and persistent deviations, while maintaining competitive detection accuracy. Future work will explore extending this region-aware compensation mechanism to multi-camera infrastructure networks and V2I cooperative perception systems to ensure broader global consistency.

We hope this work encourages future research on reliable infrastructure perception under imperfect calibration and facilitates the practical deployment of roadside vision systems in the wild.

\section{Acknowledgements}

This work was supported in part by the National Key Research and Development Program of China (No. 2023YFB4301900), by the Shenzhen Key Industry Research Program (ZDCY20250901112501002), by the Shenzhen Natural Science Foundation General Research Project (JCYJ20240813151445059), and by the Guangdong Basic and Applied Basic Research Foundation (grant No.2026A1515010917).

\bibliographystyle{splncs04}
\bibliography{main}
\end{document}